# An Ontology of Co-Creative AI Systems

Zhiyu Lin, Mark Riedl
School of Interactive Computing, College of Computing
Georgia Institute of Technology
zhiyulin@gatech.edu, riedl@cc.gatech.edu

## Abstract

The term co-creativity has been used to describe a wide variety of human-AI assemblages in which human and AI are both involved in a creative endeavor. In order to assist with disambiguating research efforts, we present an ontology of co-creative systems, focusing on how responsibilities are divided between human and AI system and the information exchanged between them. We extend Lubart's original ontology of creativity support tools with three new categories emphasizing artificial intelligence: computer-as-subcontractor, computer-as-critic, and computer-as-teammate, some of which have sub-categorizations.

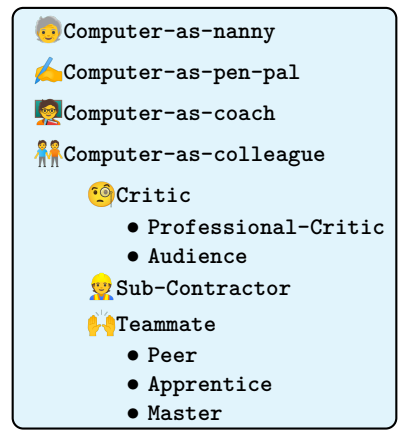

The question of whether artificial systems can be creative runs second only to the question of whether artificial systems can be intelligent, having been considered by both Ada Lovelace Menabrea [1843] and Turing [1950]. With modern, personal computing, inevitably the question was raised as whether computers can assist humans in their creative endeavors from architectural design to personal expression through art and storytelling. In 2005, Lubart [2005] introduced a special issue of the *International Journal of Human-Computer Studies* on computer support for creativity by asking how computers can be partners in the creative process. In the article, Lubart attempted to make sense of the emerging field of creativity support tools by laying out an ontology of four ways that computers can support the creative process: 👩 computer-as-nanny, ✍ computer-as-pen-pal, 👩‍🏫 computer-as-coach, and 👫 computer-as-colleague. The last category, colleague envisioned a time when artificial intelligence would be sufficiently capable to act as a partner to human creators. AI for creativity support was considered the most "ambitious" proposal, and at the time, most experiments were judged to be "failed" attempts (p. 368).

Nearly 20 years have passed and what was once considered speculative is now verging on reality. Commercial off-the-shelf text-to-image systems allow one to describe images to be generated. Increasingly, these systems use infilling to provide finer-grained control of the process. We can engage powerful, publicly-available large language models (LLMs) to write stories and poetry to our specifications, edit our own writing, or brainstorm new ideas. Music generation might be next.

However, the research is not complete. Algorithmic capabilities don't necessarily always align nicely with human-driven creative processes. There are numerous different ways in which human creators may want to engage with AI systems during creative expression. How should the responsibilities for different parts of the creative process be distributed between humans and AI? What sort of information should be exchanged between humans and AI, and when? The current popular paradigm of prompting with text is just one possibility among many Lin et al. [2023].

There is no agreed upon definition of *co-creative AI*. Whereas some use the term to refer to AI systems that possess the ability to alter the creative work equal to a human counterpart, that humans can interact with in the pursuit of their creative goals Rezwana and Maher [2021], Grabe [2022], Guzdial and Riedl [2019], there are many other ways in which one might feel they are co-creating with AI. We celebrate the diversity with which research and industry pursue co-creative systems. However, this diversity is also a challenge to researchers and practitioners alike, when a term is broadly applied to many disparate approaches.

We attempt to cast light on the different ways that computers can be "colleagues" to humans, and expand Lubart's initial ontology, based on what *responsibilities* each party is taking and the *information being exchanged*. The `computer-as-colleague` category encapsulates what is now called Co-Creative AI systems. We attempt to provide a more comprehensive—though not necessarily exhaustive—accounting for these ways. However, one will undoubtedly find that systems can span several categories. In the following paragraphs, we entail sub-categories of the `colleague` category.

💁 `Computer-as-Subcontractor` The subcontractor metaphor is arguably the most common way in which humans and AI systems interact in text-to-image generators. A subcontractor takes ownership of one constrained portion of the creation of an artifact and can operate only within the bounds of human specifications—typically the "prompt". In the case of text-to-image generators, the AI system is capable of filling a space with pixels consistent with a high-level specification. While the human gives the high-level specification, they do not intervene in the subcontractor's decisions (pixels or words) until the AI is complete. Thus, there is a delineation of agency.

🧐 `Computer-as-Critic` The critic provides feedback, or "reflections" Kreminski and Mateas [2021], but does not itself engage in alteration of the creative artifact. We delineate between two types of critics. The `Professional-Critic` provides feedback from the perspective of established norms and conventions surrounding the creative artwork (i.e., is it a good exemplar of the genre? Does it break conventions?) The `Audience` provides feedback as if it were a surrogate for the people that will experience the final creative artwork (i.e, will people like it? What will people think or feel when they experience it?). The human retains all responsibilities for altering the creative artifact in response to advice or ignoring it.

🙌 `Computer-as-Teammate` The intention of this category is for the human and the AI system to share responsibilities for the alteration of the creative artifact. Unlike the subcontractor, humans and AI have the ability to make high- or low-level decisions. Consequentially, the system should be *mixed-initiative* Novick and Sutton [1997]—human or AI can initiate changes. However, we acknowledge that not all teammates are created equal. A `Peer` relationship suggests that human and AI are approximately equally capable. Responsibilities between human and agent can be distributed in different ways: human and agent may work on the same aspects of the artifact at the same time, or different aspects of the artifact can be partitioned off for human and agent. An `Apprentice` relationship suggests the human has greater capabilities than the AI. The AI could be intentionally given fewer capabilities , or the technology could not be (yet) at human-parity. A `Master` relationship (the reciprocal of an apprentice) suggests the AI has capabilities superior to the human partner. Open questions include: knowing when it is appropriate for the agent to take initiative, how to adapt to the human's process, and how to augment and extend the human creator's abilities.

💆 `Computer-as-Coach` While coaching sits outside the "colleague" category, there are also opportunities for AI innovation here. A coach is an entity that does not engage directly in the human activity, as in sports, as well as the creation of an artifact; the coach helps the human user learn and refine their skills. This can be done without AI, such as using a fixed curriculum of practice. AI may augment the initial vision for the coach, for example by learning a model of a user's skills and generating personalized tutorials. To the extent that a coach might also provide advice, there is overlap with the `Master` if a coach provides actionable advice at the moment; however, the key distinguisher would be the ability to directly contribute to the creative artifact.

**Conclusions** We have shown the history of co-creativity and its modern diversification. We present an ontology beyond what Lubart proposed on co-creative systems, focusing on the responsibilities of different parties in the co-creative activity, and the information exchanged between them, and sketched outlooks of these systems. We encourage designers of future co-creative systems to utilize this ontology to cast understanding on the co-creative aspect of their system, and further use it to discover opportunities in their designs beyond the creative capability of the AI agents alone.

## Ethical Statement

Although research on co-creativity related to this generation of AI and Machine Learning techniques is emerging, most emphasizes capability of these models as opposed to the human-facing aspects of it. We argue that a lack of research in this human-AI assemblage is the most major ethical concern. As the nature of co-creative systems is to interact with human users and enhance their potential, additional concerns arise beyond what matters for a question-answering system. Ehsan et al. Ehsan and Riedl [2021] pointed out that what agents present to the user may not be what they believe internally, and careless handling of these interactions is a threat to the users, in decision-making systems; Rezwana et al. Rezwana and Maher [2022] signals the same threat in the domain of co-creativity by discovering that a co-creative agent that expresses human-like traits is deemed "more reliable", regardless of the actual creative effort of the agent; Buschek et al. Buschek et al. [2021] hinted that aside from common ethical concerns of AI agents, how much AI is contributing to the human-AI assemblage should also be under scrutiny. Since ideas shared by the AI systems can steer the creative process, causing the same bias a decision-making agent can introduce, we encourage researchers alike to conduct further studies into co-creative systems, especially beyond the generative capability of their AI system. By disambiguating different ways in which human-AI assemblages can form in creative domains, we allow for more fine-grained analysis of the potential ethical considerations that can arise from co-creative AI systems.